\documentclass[letterpaper, 10 pt, conference]{ieeeconf}  

\IEEEoverridecommandlockouts                              

\makeatletter
\let\NAT@parse\undefined
\makeatother
\usepackage[numbers]{natbib}

\usepackage{graphics} 
\usepackage{amsmath} 
\usepackage{amssymb}  
\usepackage{xcolor}
\usepackage{cleveref}
\usepackage{siunitx}
\usepackage{graphicx}
\usepackage{mathtools}
\usepackage{subcaption}
\usepackage{multirow}
\usepackage{url}

\title{\LARGE \bf
Walk like Dogs: Learning Steerable Imitation Controllers \\
for Legged Robots from Unlabeled Motion Data
}

\newcommand{\crl}{\ast}
\newcommand{\ril}{\dagger}

\author{Dongho Kang$^{\crl}$, Jin Cheng$^{\crl}$, Fatemeh Zargarbashi$^{\crl}$, Taerim Yoon$^{\ril}$, Sungjoon Choi$^{\ril}$, and Stelian Coros$^{\crl}$
\thanks{This work has received funding from the European Research Council (ERC) under the European Union’s Horizon 2020 research and innovation programme (grant agreement No. 866480.)}
\thanks{$^{\crl}$The authors are with the Computational Robotics Lab in the Department of Computer Science, ETH Zurich, Switzerland.
{\tt\footnotesize \{kangd, jicheng, fzargarbashi, scoros\}@ethz.ch}}%
\thanks{$^{\ril}$The authors are with the Robot Intelligence Lab in the Department of Artificial Intelligence, Korea University, South Korea.
{\tt\footnotesize \{taerimyoon, sungjoon-choi\}@korea.ac.kr}}%
}

\begin{document}

\maketitle
\thispagestyle{empty}
\pagestyle{empty}

\newcommand{\DK}[1]{{\bf\textcolor{red}{DK: #1}}}
\newcommand{\JC}[1]{{\bf\textcolor{blue}{JC: #1}}}
\newcommand{\FZ}[1]{{\bf\textcolor{orange}{FZ: #1}}}
\newcommand{\TY}[1]{{\bf\textcolor{magenta}{TY: #1}}}
\newcommand{\Diff}[1]{{\textcolor{black}{#1}}}

\newcommand{\norm}[1]{\left\lVert#1\right\rVert}

\newcommand{\Rthree}{\mathbb{R}^3}
\newcommand{\SOthree}{SO(3)}

\newcommand{\worldframe}{\mathcal{W}}
\newcommand{\baseframe}{\mathcal{B}}
\newcommand{\groundprojectedframe}{\mathcal{P}}

%
\newcommand{\state}{\mathbf{x}}
\newcommand{\control}{\mathbf{u}}
\newcommand{\pos}{\mathbf{r}}
\newcommand{\euler}{\mathbf{\Theta}}
\newcommand{\quat}{\mathbf{h}}
\newcommand{\limb}{\mathbf{e}}
\newcommand{\gencoord}{\mathbf{q}}
\newcommand{\genvel}{\dot{\mathbf{q}}}
\newcommand{\genacc}{\ddot{\mathbf{q}}}
\newcommand{\sourcemotion}{{}^{\text{src}}}
\newcommand{\possource}{\sourcemotion\mathbf{r}}
\newcommand{\eulersource}{\sourcemotion\mathbf{\Theta}}
\newcommand{\quatsource}{\sourcemotion\mathbf{h}}
\newcommand{\limbsource}{{}_{\mathcal{B}}^{\text{src}}\mathbf{e}}
\newcommand{\possourcelocal}{{}_{\baseframe}^{\text{src}}\mathbf{r}}
\newcommand{\targetmotion}{{}^{\text{tgt}}}
\newcommand{\postarget}{\targetmotion\mathbf{r}}
\newcommand{\quattarget}{\targetmotion\mathbf{h}}
\newcommand{\postargetlocal}{{}_{\baseframe}^{\text{tgt}}\mathbf{r}}
\newcommand{\kinmotion}{{}^{\text{kin}}}
\newcommand{\refmotion}{{}^{\text{ref}}}
\newcommand{\joystick}{\mathbf{c}}

\newcommand{\vaemotion}{{}^{\text{vae}}}
\newcommand{\gp}{{}_{\groundprojectedframe}}

%
\newcommand{\rewardi}{r_I}
\newcommand{\weighti}{w_I}
\newcommand{\rewardw}{r_{\mathcal{W}}}
\newcommand{\weightw}{w_{\mathcal{W}}}
\newcommand{\rewardr}{r_R}
\newcommand{\weightr}{w_R}

\begin{abstract}


\Diff{We present an imitation learning framework that extracts distinctive legged locomotion behaviors and transitions between them from unlabeled real-world motion data. By automatically discovering behavioral modes and mapping user steering commands to them, the framework enables user-steerable and stylistically consistent motion imitation.}
Our approach first bridges the morphological and physical gap between the motion source and the robot by transforming raw data into a physically consistent, robot-compatible dataset using a kino-dynamic motion retargeting strategy. This data is used to train a steerable motion synthesis module that generates stylistic, multi-modal kinematic targets from high-level user commands. These targets serve as a reference for a reinforcement learning controller, \Diff{which reliably executes them on the robot hardware.
In our experiments, a controller trained on dog motion data demonstrated distinctive quadrupedal gait patterns and emergent gait transitions in response to varying velocity commands. These behaviors were achieved without manual labeling, predefined mode counts, or explicit switching rules, maintaining the stylistic coherence of the data.}

\end{abstract}

\section{Introduction}



Imitation learning offers an efficient way to acquire stylistic and versatile skills for legged robots by leveraging real-world locomotion data \cite{peng2020learning, yoon2024spatio, liao2025beyondmimic}.
By directly imitating motion data, this approach eliminates the need for hand-crafted rules or heuristic control objectives.

\Diff{
However, significant challenges remain. First, morphological and physical discrepancies between the motion source and target robot often impede high-fidelity imitation. Second, to move beyond fixed trajectory playback, the controller must be capable of responding to interactive user commands. Third, preserving the diversity of the original motion data is essential; this behavioral richness allows the robot to map steering commands to appropriate behavioral modes, thereby enhancing responsiveness.
}


To address these challenges, this paper presents a framework for steerable imitation in legged robots by leveraging the diverse movement patterns present in an \Diff{unlabeled} real-world motion database (DB).
\Diff{In this context, \emph{steerable} refers to resulting controller's ability to autonomously switch between different behavioral modes to match high-level user commands. The term \emph{unlabeled} indicates that our approach automatically identifies these behavioral modes from a rich dataset containing a wide range of locomotion patterns spanning various movement speeds, notably without the data being pre-segmented or pre-labeled.}

The framework begins with kino-dynamic motion retargeting, which adapts the motion data to the robot's morphology and physical capabilities, mitigating kinematic and physical artifacts. 
To generate steerable behaviors, we employ a variational autoencoder (VAE)-based motion synthesis module. This module produces reference robot motions conditioned on user steering commands, capturing distinct behavioral modalities while preserving the stylistic consistency and inherent diversity of the unlabeled data.
These reference motions are executed by a feedback control policy, trained via RL with the motion synthesis module in the loop.

\begin{figure}[!t]
    \centering
    \includegraphics[width=\linewidth]{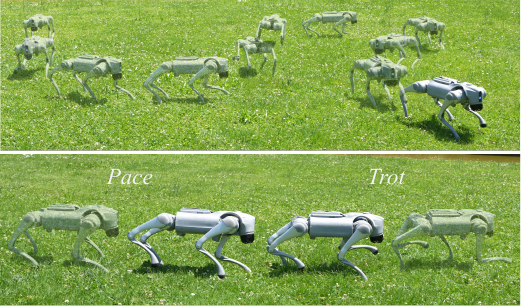}
    \caption{\emph{Unitree Go2} robot navigating freely across a grass field in response to joystick commands (\textbf{top}). The gait pattern automatically transitions from \emph{Pace} to \emph{Trot} as the forward speed command increases from \SI{0.6}{\metre/\second} to \SI{1.0}{\metre/\second} (\textbf{bottom}).}
    \label{fig:hw-experiment-online}
    \vspace{-2em}
\end{figure}


\Diff{We demonstrate the capability of our framework through a series experiment in quadrupedal locomotion learned from dog motion data.}
In our experiments, we verify that our kino-dynamic motion retargeting method effectively generates kinematically and dynamically feasible motions, enabling reliable RL training that accurately replicates dynamic and expressive skills. 
By effectively preserving multiple gait modes from the motion database, the motion synthesis module generates reference trajectories that not only track user-specified velocity commands accurately but also autonomously switch between gait patterns to enable responsive steering. 
Finally, we demonstrated the full pipeline by deploying the motion synthesis and control modules online on the \emph{Unitree Go2} robot, showcasing user-steerable locomotion and gait-switching behavior in real time, as shown in \Cref{fig:hw-experiment-online}.

\section{Related Work}

\subsection{Real-world Motion Data Imitation in Legged Robotics}


To replicate the natural and versatile movements of humans and animals in robotic systems, a growing body of research leverages motion data captured from real subjects. 

A first challenge in this direction is bridging the morphological and physical gap between the motion source and the target robot---a process known as \emph{motion retargeting} (MR). 
Earlier efforts primarily addressed kinematic discrepancies by using scaled keypoint transfers \cite{choi2019towards}, or remapping high-level motion features such as contact timings and base trajectories \cite{kang2021animal, kang2022animal}.
However, when the source motion exhibits complex and highly dynamic behaviors, difference in physical capabilities between the source and the robot become increasingly critical. 
To address this, dynamic MR methods that account for the robot’s dynamics and physical limits have gained increasing attention. These methods commonly employ model-based control frameworks \cite{yoon2024spatio, zhang2023slomo, grandia2023doc} as offline optimization tools to refine reference motions, ensuring dynamic feasibility with the target robot.

A second major challenge lies in developing a robust feedback controller capable of executing these motions on hardware while preserving their inherent agility and expressiveness.
In this context, imitation learning via reinforcement learning (RL) is increasingly favored for its ability to produce robust control policies across diverse motion repertoires \cite{peng2020learning, escontrela2022adversarial, tang2024humanmimic, zargarbashi2024robot, liao2025beyondmimic}, while overcoming the runtime computational demands and modeling limitations of the classical model-based control approaches.



Our approach incorporates a kino-dynamic MR strategy that addresses both the kinematic and physical gap between the source and the target robot. Using constrained inverse kinematics and model-based control, it generates a robot motion DB consisting of kinematically and dynamically feasible sequences. 
To achieve real-time motion tracking, we employ RL to train a residual motion tracking policy that reliably executes reference motions on the physical robot.  

\begin{figure*}
    \vspace{2.5mm}
    \centering
    \includegraphics[width=0.95\linewidth]{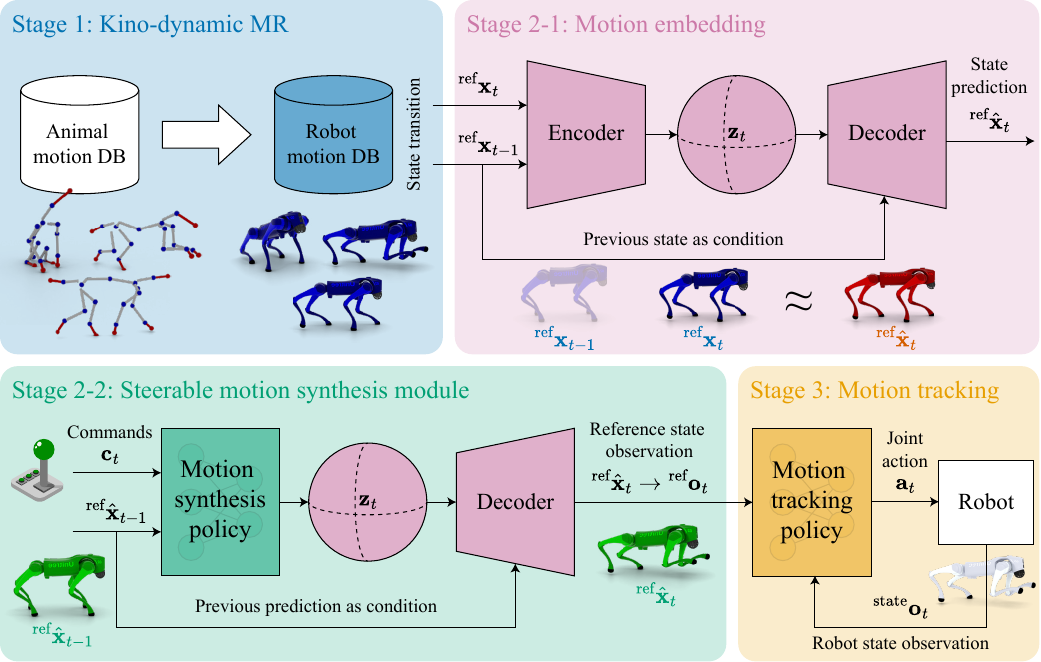}
    \caption{Overview of the framework. An animal motion DB is first transformed into a robot motion DB using kino-dynamic motion retargeting (in \textbf{blue}). Next, each state transition in the motion DB is embedded into a latent space using a VAE (in \textbf{purple}). The trained decoder, combined with an RL-based motion synthesis policy produces a new reference motion in response to steering commands (in \textbf{green}). Finally, the reference motion is tracked by an RL controller (in \textbf{orange}).}
    \label{fig:overview}
    \vspace{-1em}
\end{figure*}

\subsection{Steerable and Stylistic Motion Synthesis}

The field of character animation has widely explored the steerable motion synthesis using real-world data, aiming to create motions that are realistic, diverse, and interactive. 
Recently, this goal has driven a significant shift towards learning-based generative methods, particularly those using generative adversarial networks (GANs)~\cite{peng2021amp, peng2022ase}, VAEs~\cite{ling2020character, yao2022controlvae}, and more recently, diffusion models~\cite{truong2024pdp, huang2025diffuse}.



Notably, GAN-based approaches have gained attention for generating steerable behaviors in the context of physics-based characters.
Adversarial Motion Priors (AMP)~\cite{peng2021amp} is a prominent example, wherein an RL policy acts as the \emph{generator} to synthesize natural motions in a physics simulation, while a \emph{discriminator} guides the process by rewarding stylistic realism.
When combined with task rewards, this method enables steerable behaviors while retaining realistic motion style.
\Diff{However, AMP and its variant are prone to training instability and mode collapse, which can limit motion diversity and require extensive hyperparameter tuning.
Furthermore, their end-to-end nature lacks the transparency needed to easily diagnose and correct errors.}

\Diff{To address these limitations, the research community is increasingly exploring diffusion-based approaches \cite{truong2024pdp, huang2025diffuse}. Diffusion models inherently excel at capturing complex data distributions and effectively preserving diversity within motion datasets. However, their prohibitive runtime computational cost typically precludes real-time execution on resource-constrained hardware.}

Another line of research leverages VAEs to embed motion datasets into a structured latent space for steerable motion generation.
\citet{ling2020character, won2022physics} proposed VAE architectures where state transitions are embedded in the latent space as conditional distributions.  
In these frameworks, a decoder generates the next state prediction based on the previous state and a latent vector, effectively capturing motion dynamics. 
To enable control, an RL policy modulates these latent vectors, ensuring the resulting motion transitions align with user commands while preserving motion style.



Our framework builds upon the approaches that combine a VAE and an RL policy~\cite{ling2020character, won2022physics}. Our key distinction is the introduction of a hyperspherical latent space~\cite{peng2022ase, davidson2018hyperspherical}. This provides the RL policy with a well-defined action space that prevents unbounded exploration, which is critical for preserving stylistic coherence and diversity.
\section{Overview}

An overview of our framework is illustrated in \Cref{fig:overview}. 
\Diff{Our framework consists of three stages: kino-dynamic motion retargeting, the development of a steerable motion synthesis module, and the training of an RL tracking controller for physical robot deployment.}

As a first step, we transform \Diff{unlabeled} real-world motion dataset into a robot-compatible motion DB. 
Specifically, we use constrained inverse kinematics and a model-based control framework to ensures the resulting motion sequences are both kinematically and dynamically feasible for the robot.

\Diff{
Subsequently, we develop steerable motion synthesis module based on a \emph{hyperspherical} VAE \cite{davidson2018hyperspherical}. 
The VAE embeds the state transitions from the retargeted motion DB into a latent space by learning to reconstruct the current state $\refmotion\hat{\state}_{t}$, conditioned on the previous state $\refmotion\state_{t-1}$ and a latent vector $\mathbf{z}_t$. 
Once this embedding is established, a reference motion synthesis module is developed via RL to map the user steering command $\joystick_t$ (in our implementation, body velocity commands) to corresponding reference motions, while maintaining stylistic coherence with the motion DB. 
Specifically, the motion synthesis policy modulates latent variables conditioned on $\joystick_t$ and the previously generated reference motion $\refmotion\hat{\state}_{t-1}$.
The policy learns to navigate the latent space, producing new reference motion $\refmotion\hat{\state}_{t}$ that effectively tracks arbitrarily steering commands during training.
Crucially, the hyperspherical latent space ensures that a latent variable $\mathbf{z}_t$ remains bounded, providing a well-defined action space for the motion synthesis policy. 
This design choice is essential for maintaining stylistic consistency with the data and preserving behavioral diversity.
}

Finally, the synthesized reference motion $\refmotion\hat\state_t$ is provided as a tracking target to an RL feedback tracking controller, trained to robustly track the reference motion on a physical robot.
At runtime, we deploy the pipeline comprising the motion synthesis policy, the VAE decoder, and the RL tracking controller to enable the robot to interactively respond to user commands.
The subsequent chapters describe each stage of this framework in detail. 

\section{Kino-dynamic Motion Retargeting}
\label{sec:motion-retargeting}

We use a \emph{kino-dynamic} MR approach that combines constrained inverse kinematics (IK) with a model-based control framework. This ensures that motions retargeted from real-world motion are both kinematically and dynamically feasible for the robot.
Similar to the previous work by \citet{yoon2024spatio}, we decouple the MR process into two stages: a kinematics stage and a dynamics stage.

In the kinematics stage, we process a source animal's motion pose-by-pose starting from the first time instance. 
At each time step, we extract the source base position $[{\sourcemotion}x, {\sourcemotion}y, {\sourcemotion}z]$, base roll, pitch and yaw angles $[{\sourcemotion}\phi, {\sourcemotion}\theta, {\sourcemotion}\psi]$, and limb vectors $\limbsource_i$ with $i \in \{1, 2, 3, 4\}$. The limb vectors are unit vectors, expressed in the base frame $\{\baseframe\}$, pointing from the shoulders to the corresponding foot.
Additionally, we compute the ground-projected forward velocity, ${\sourcemotion}v_{\text{fwd}}$, sideway velocities ${\sourcemotion}v_{\text{side}}$, and yaw rate ${\sourcemotion}\dot{\psi}$ of the source; these quantities are used as 2D velocity components for subsequent processing.

We adjust the base height, roll, and pitch components by applying scaling factors $\alpha_{(\cdot)} \in \mathbb{R}$, and transfer to the robot:
\begin{equation}
\begin{bmatrix}
{\targetmotion}z &
{\targetmotion}\phi &
{\targetmotion}\theta 
\end{bmatrix}
= 
\begin{bmatrix}
\alpha_z \, {\sourcemotion}z &
\alpha_{\phi} \, {\sourcemotion}\phi &
\alpha_{\theta} \, {\sourcemotion}\theta
\end{bmatrix}.    
\end{equation}
Additionally, we scale and numerically integrate the 2D velocity components for the remaining base pose components:
\begin{equation}
\begin{bmatrix}
{\targetmotion}x \\
{\targetmotion}y \\
{\targetmotion}\psi \\
\end{bmatrix}
= 
\begin{bmatrix}
{\targetmotion}x^{-} \\
{\targetmotion}y^{-} \\
{\targetmotion}\psi^{-} \\
\end{bmatrix} + \Delta t
\mathbf{R}_z({\targetmotion}\psi^{-})
\begin{bmatrix}
\alpha_{\text{fwd}} \,  {\targetmotion}v_{\text{fwd}}^{-} \\
\alpha_{\text{side}} \, {\targetmotion}v_{\text{side}}^{-} \\
\alpha_{\dot{\psi}} \,  {\targetmotion}\dot{\psi}^{-} \\
\end{bmatrix},
\end{equation}
where the superscript ${(\cdot)}^{-}$ denotes the value at the previous timestep, and $\mathbf{R}_z(\cdot) \in \mathbb{R}^{3\times3}$ is the rotation matrix about $z$-axis. Subsequently, we compute the robot's foot positions by scaling limb vectors with the factor $\boldsymbol{\alpha}_{\text{limb}} \in \Rthree$ and adding them to the shoulder positions:
\begin{equation}
\postarget_{\text{foot},i}  = \postarget_{\text{shoulder},i} + \targetmotion\mathbf{R}_{\mathcal{WB}} (\boldsymbol{\alpha}_{\text{limb}} \odot \limbsource_i),
\end{equation}
where $\odot$ denotes element-wise multiplication and $\targetmotion\mathbf{R}_{\mathcal{WB}}$ is the rotational matrix of the robot base. The values of the scaling factors used in our experiments are listed in \Cref{tab:retargeting-kin-hyperparams}. 

\begin{figure}[!t]
    \vspace{2mm}
    \centering
    \includegraphics[width=0.95\linewidth]{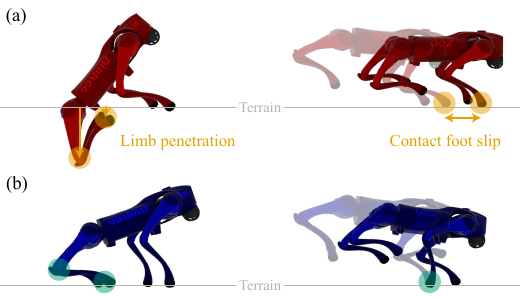}
    \caption{(\textbf{a}) Limb penetration and contact foot slips introduced by UVM. (\textbf{b}) Our kino-dynamic MR removes these artifacts, and ensure both kinematic and dynamic feasibility.}
    \label{fig:motion-retargeting}
    \vspace{-1em}
\end{figure}

After constructing base pose and foot positions, we can use a standard IK solver to compute generalized coordinates of the robot $\targetmotion\gencoord$ which allow us to build full pose of the robot \cite{choi2019towards}. However, this scale-and-transfer method---often refer to as unit vector method (UVM)---introduces kinematic artifacts such as contact foot slips, limb penetrations, and violation of joint limits as illustrated in \Cref{fig:motion-retargeting}.



To address these issues, we formulate a \emph{constrained} IK problem that enforces limb and joint constraints as follows: 
\begin{subequations}
\label{eq:ik}
\begin{alignat}{3}
    &\min_{\gencoord}\ 
    \Vert && {\targetmotion}\gencoord_{\text{base}} \ominus \gencoord_{\text{base}} \Vert^2
    + \sum_{k \in \text{swing}} \Vert \postarget&&_{\text{foot},k} - \text{FK}_{\text{foot},k}(\gencoord) \Vert^2 \nonumber \\
    & \:\: \text{s.t.} 
    && \text{FK}_{\text{foot},j}(\gencoord) = \pos_{\text{anc},j}     
    && \forall j \in \text{stance}   \label{eq:ik-stance-const} \\
    &
    && \text{FK}_{\text{foot},k}(\gencoord)_z > 0
    && \forall k \in \text{swing}    \label{eq:ik-swing-const}  \\
    &
    && \text{FK}_{\text{knee},i}(\gencoord)_z > 0
    && \forall i \in \{1, 2, 3, 4\}  \label{eq:ik-knee-const} \\
    &
    && \bar{\gencoord} > \boldsymbol{\gencoord} > \underline{\gencoord}. 
    && \label{eq:ik-joint-const} 
\end{alignat}    
\end{subequations}
Here, $\text{FK}_{\text{foot},i}(\cdot)$ and $\text{FK}_{\text{knee},i}(\cdot)$ are forward kinematics functions mapping the robot's generalized coordinates to the world-frame position of the $i$-th foot and knee, respectively. 
The operator $\ominus$ denotes the substraction between base pose components of generalized coordinates, performing direct subtraction for positions and computing a scalar error from the quaternion difference for orientation. 

The objective function of \Cref{eq:ik} consists of two terms: 1) the error between the base pose components of generalized coordinates variable ${\targetmotion}\gencoord_{\text{base}}$ and its IK target $\gencoord_{\text{base}}$, and 2) the positional error between the swing feet and their respective IK targets.
Importantly, the constraint \Cref{eq:ik-stance-const} fixes the position of each stance foot $j$ to an anchor position $\pos_{\text{anc},j}$, where the $z$-component is set to the terrain height, and $x-$ and $y-$components are taken from the foot's position when the current contact was initially detected. This constraint prevents footslip and foot penetration in the retargeted motion. 
Additionally, \Cref{eq:ik-swing-const} ensures that the swing foot is always above the terrain. 
Simiarly, \Cref{eq:ik-knee-const} ensures that every knee is always above the terrain.
Finally, \Cref{eq:ik-joint-const} ensures that the generalized coordinates respect their limits, such as the joint limits. 

The optimal solution from this optimization is used as a kinematically retargeted motion $\kinmotion\gencoord$, and processing the entire source motion frame by frame yields the full retargeted motion sequence $\kinmotion\gencoord_{1:N}$ where $N$ is the sequence length.

After obtaining $\kinmotion\gencoord_{1:N}$, we proceed to the dynamics stage to ensure dynamic feasibility using a model-based control framework---specifically, model predictive control (MPC).
At a high-level, this process is formulated as a trajectory tracking problem, where the objective is to follow $\kinmotion\gencoord_{1:N}$ by minimizing the generalized coordinate error over a time horizon $T$ in a receding horizon manner.
The control input is the robot's joint torque, bound by the robot's actuation limits, and the system state $\state$ contains the generalized coordinates and velocities: $\state \coloneqq [\gencoord, \genvel]$.

\begin{table}
    \vspace{2mm}
    \caption{Default scaling factors used in the MR kinematic stage in our experiments.}
    \label{tab:retargeting-kin-hyperparams}
    \vspace{-0.5em}
    \begin{center}
    \begin{tabular}{|c|c|c|c|c|c|c|}
    \hline
    $\alpha_z$              & $\alpha_\phi$             & $\alpha_\theta$       &
    $\alpha_{\text{fwd}}$   & $\alpha_{\text{side}}$    & $\alpha_{\dot{\psi}}$ &
    $\boldsymbol{\alpha}_{\text{limb}}$ 
    \\ \hline 
    $0.81$                  & $1.0$                     & $1.0$                 &
    $0.6$                   & $0.6$                     & $1.0$                 &
    $[0.6, 0.7, 0.81]$ 
    \\ \hline
    \end{tabular}
    \end{center}
    \vspace{-2em}
\end{table}

We implement this MPC using the \emph{MJPC} framework \cite{howell2022predictive}, with the \emph{iLQG} solver \cite{tassa2012synthesis}, which leverages the full-body robot model and contact model of the \emph{MuJoCo} simulator \cite{todorov2012mujoco} to numerically compute the derivative information required by the solver. 
In our experiment, we set the time horizon to $T=\SI{2.0}{\second}$ with a discretization time step of $\SI{0.01}{\second}$.

The resulting MPC trajectory serves as our kino-dynamically retargeted motion $\refmotion\state_{0:N}$ and is stored in our retargeted motion DB. 
Although solving this MPC problem is computationally expensive, MR is performed offline to generate a robot motion database for downstream learning tasks, which does not require real-time execution.

\section{Steerable Motion Synthesis}
\label{sec:motion-synthesis}


The steerable motion synthesis module is a central component of our framework, responsible for preserving the stylistic quality of motions in the DB, while capturing and reproducing their diversity in response to steering commands. 
We adopt a VAE-based motion synthesis approach which effectively preserves the behavioral multi-modality in the motion data. 
This approach comprises two substages: motion embedding and the RL training of a motion synthesis policy.

\subsection{Motion Embedding}

Firstly, we embed state transitions present in the motion DB into a structured latent space, by training a VAE to reconstruct the latter state in each transition pair.
The input and output of the VAE are defined using local components of the robot state. 
Specifically, we introduce a ground-projected base frame $\{\groundprojectedframe\}$, and define the VAE state vector as
\begin{equation}
    \vaemotion\state \coloneqq
    \begin{bmatrix} 
    {\gp}z                          &
    \gp\mathbf{h}                   &  
    {\gp}\mathbf{v}                 &
    {\gp}\mathbf{w}                 &
    {\gp}\pos_{\text{feet}}         &
    \boldsymbol{\theta}             &
    \dot{\boldsymbol{\theta}}       
    \end{bmatrix} \in \mathbb{R}^{49},
\end{equation}
which concatenates the base height, base orientation, base linear and angular velocities, foot positions relative to ${\groundprojectedframe}$, joint angles, and joint speeds. The base orientation $\gp\mathbf{h} \in \mathbb{R}^6$ is represented using the $x-$ and $z-$axis vectors of the base frame, expressed in $\{\groundprojectedframe\}$ \cite{zhou2019on}.

The VAE encoder takes as input a pair of consecutive states $(\vaemotion\state_{t-1}, \vaemotion\state_{t})$, and outputs distribution parameters of the 18-dimensional latent space. 
The decoder then predicts the current state $\vaemotion\hat\state_{t}$ conditioned on the latent vector $\mathbf{z}_t$ and the previous state $\vaemotion\state_{t-1}$. 
We adopt a mixture of experts (MoE) architecture for the decoder, consisting of six expert networks and a gating network similar to the previous work by \citet{ling2020character}. 
The encoder and each expert network in the decoder are implemented as fully connected networks with two hidden layers of 256 units. The gating network is also a fully connected network, with two hidden layers of 64 units. All hidden layers use the ELU activation function.

To structure the latent space, we use a hyperspherical latent representation \cite{davidson2018hyperspherical} by modeling the latent variable distribution as a von Mises-Fisher (vMF) distribution instead of a standard Gaussian distribution. 
This design choice is crucial for streamlining the training of the motion synthesis policy in the next stage, as it constrains the action space to the bounded surface of a hypersphere \cite{peng2022ase}. 
Without this measure, the synthesis module is prone to losing stylistic coherence or failing to reproduce key gait modes from the data.
The VAE training loss is defined as:
\begin{equation}
    \mathcal{L} = \Vert \vaemotion\state_{t} - \vaemotion\hat\state_{t} \Vert^2_2 
    + \beta D_{\text{KL}} \left( q \, \Vert \, p \right),
\end{equation}
where $q(\mathbf{z}_t\vert\vaemotion\state_{t}, \vaemotion\state_{t-1})$ is the posterior distribution approximated by the encoder and $p(\mathbf{z}_t)$ is the vMF prior.
We refer readers to the work by \citet{davidson2018hyperspherical} for the derivation of KL divergence term for vMF distribution. 
In our experiments, we set the weighting coefficient to $\beta = 0.05$.

We initially train the VAE using state transitions $(\vaemotion\state_{t-1}, \vaemotion\state_{t})$ from the motion DB for 20 epochs. We then gradually shift to autoregressive training over 60 epochs, where the decoder's prediction $\vaemotion\hat\state_{t}$ is recursively used as the next input condition (i.e. $\vaemotion\hat\state_{t} \rightarrow \vaemotion\state_{t-1}$). This training strategy improves the stability of sequence prediction in downstream tasks.

\begin{table}
    \vspace{2mm}
    \caption{PPO hyperparameters.}
    \label{tab:motion-synthesis-policy-hyperparams}
    \vspace{-0.5em}
    \begin{center}
    \begin{tabular}{|l|c||l|c|}
    \hline
    Number of envs      &   $4096$      & Value ftn. coeff.  &  $1.0$   \\ \hline
    Batch size          &   $24576$     & Entropy coeff.$^*$ &  $0.002$   \\ \hline
    Number of epochs    &   $5$         & Discount factor    &  $0.99$  \\ \hline
    Learning rate       &   $0.0005$    & GAE parameter      &  $0.95$  \\ \hline      
    NN Hidden layers    &   $[512, 256, 128]$   & Clipping range    & $0.2$ \\ \hline
    NN Activation ftn.  & ELU                   & KL target         & $0.01$\\ \hline
    \end{tabular}
    \caption*{\scriptsize{$^{*}$ We set the entropy coefficient to $0.0$ for the motion synthesis policy.}}
    \end{center}
    \vspace{-3em}
\end{table}

\subsection{Motion Synthesis Policy}

In the second stage, we train the motion synthesis policy using RL to navigate the hyperspherical latent space constructed in the previous stage and generate states that follow the user's steering commands.  

The policy observes the user's forward and turning speed commands $\joystick_t \coloneqq [c_{\text{fwd}, t}, c_{\text{turn}, t}]$ and the previously generated motion state $\vaemotion\hat{\state}_{t-1}$. 
The action is defined as a vector $\tilde{\mathbf{z}}_t \in \mathbb{R}^{18}$ which is later projected onto the hyperspherical latent space by $\mathbf{z}_t = \tilde{\mathbf{z}}_t / \Vert \tilde{\mathbf{z}}_t \Vert_2$.

The policy is trained with Proximal Policy Optimization (PPO) algorithm  \cite{schulman2017proximal} using a reward function designed to align the ground-projected forward speed of the robot's base $v_{\text{fwd}}$ and yaw rate $\dot{\psi}$ with their commanded values:
\begin{equation}
    r = 
    \exp \left( 
    -\frac{(v_{\text{fwd}} - c_{\text{fwd}})^2}{0.25} 
    -\frac{(\dot{\psi} - c_{\text{turn}})^2}{0.1} \right) 
\end{equation}

The trained RL policy, together with the decoder, constitutes the motion synthesis module, which generates reference motions in real time in response to user commands.
The list of the PPO hyperparameters used is provided in \Cref{tab:motion-synthesis-policy-hyperparams}.

\section{Motion Tracking via Residual Policy}


As a final step, we use RL to train a policy that tracks a synthesized reference motion robustly on physical robot.  
The policy is designed to generate residual joint positions that are added to the reference motion, enabling the system to compensate for the discrepancies between the synthesized motion and real-world dynamics.

\begin{figure*}
    \vspace{2.5mm}
    \centering
    \includegraphics[width=0.95\linewidth]{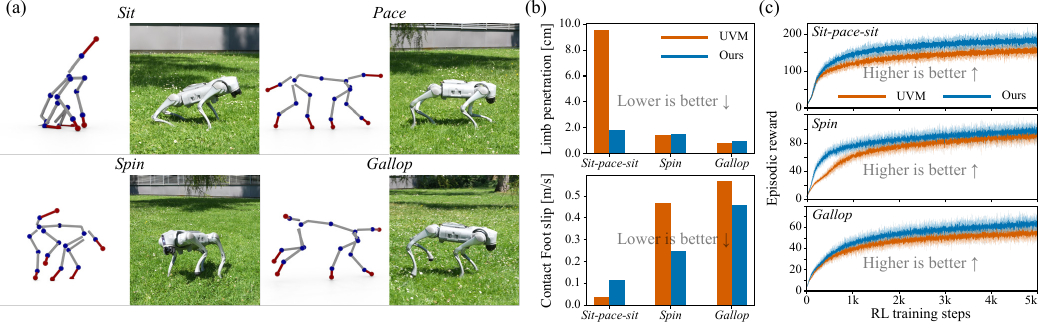}
    \caption{(\textbf{a}) Our kino-dynamic MR enables reliable transfer of dog motion sequences to the real-world robot. We evaluate its effectiveness against the UVM baseline by comparing (\textbf{b}) kinematic artifacts in the retargeted motions and (\textbf{c}) the resulting RL training curves for downstream imitation tasks, where RL policies are trained to execute the motions.}
    \label{fig:results-motion-retargeting}
    \vspace{-1em}
\end{figure*}

\begin{table}
    \vspace{2mm}
    \caption{Motion tracking policy reward hyperparameters.}
    \label{tab:reward-hyperparams}
    \vspace{-0.5em}
    \begin{center}
    \begin{tabular}{|l|c|}
    \hline
    Reward terms $r_{\mathrm{y}}$          & Sensitivity $\sigma_\mathrm{y}$ \\ \hline
    Base linear velocity $r_\mathrm{v}$         & $0.2$   \\ \hline
    Base angular velocity $r_\mathrm{w}$        & $0.25$  \\ \hline
    Base height $r_z$                           & $0.1$   \\ \hline
    Base orientation $r_{\phi\theta}$           & $0.8$   \\ \hline
    Feet position $r_{\text{feet}}$             & $[0.3, 0.3, 0.1]^{*}$ \\ \hline
    Global position $r_{xy}$                    & $0.5$   \\ \hline
    Global orientation $r_{\mathrm{h}}$         & $0.5$   \\ \hline
    Action rate $r_{\Delta\mathrm{a}}$          & $2.0$   \\ \hline
    Action scale $r_{\mathrm{a}}$               & $10.0$  \\ \hline
    Feet slip $r_{\text{slip}}$                 & $0.1$   \\ \hline
    \end{tabular}
    \caption*{\scriptsize{
    $^{*}$ Non-scalar values are applied element-wise.
    }}
    \vspace{-3em}
    \end{center}
\end{table}

\subsection{Observation and Action Space}

At timestep $t$, the policy (actor) receives three sets of inputs: the first one is a noisy observation of robot's state ${}^{\text{state}}\mathbf{o}_t \in \mathbb{R}^{42}$, which includes the gravity vector, angular velocity, joint angles, joint speeds, and previous joint commands; the second one is a reference observation ${}^{\text{ref}}\mathbf{o}_t \in \mathbb{R}^{26}$, which provides the target height, gravity vector, linear and angular velocities, joint angles, and foot heights. All velocity and gravity vectors are expressed in the base frame. 
The full observation vector is formed as a stack of the state and reference motion observation with their history, with the history length $H=4$. Additionally, the latent vector $\mathbf{z}_t \in \mathbb{R}^{18}$ from the motion synthesis module is included:
\begin{equation}
\mathbf{o}_t = 
\begin{bmatrix}
{}^{\text{state}}\mathbf{o}_{t-H:t} & 
{}^{\text{ref}}\mathbf{o}_{t-H:t} &
\mathbf{z}_t
\end{bmatrix} \in \mathbb{R}^{358}. 
\end{equation}
As animal motions involve frequent flying phases and irregular stepping patterns, estimating the robot's base height and linear velocity becomes challenging. Therefore, we chose to exclude these components from the observation.

The critic takes as input the noise-free version of the policy observations, along with privileged information including the base height, linear velocity, actual and reference foot positions relative to the base, and foot velocities. 
Additionally, it receives the position and orientation error of the base in the world frame to encourage exploration for dynamic motions. 

The output of the policy is residual joint actions $\mathbf{a} \in \mathbb{R}^{12}$, which is added to the reference joint angles ${}^{\text{ref}}\boldsymbol{\theta}_t$ with a scaling factor and set as the PD targets of the robot joints, ${}^{\text{PD}}\boldsymbol{\theta}_t = {}^{\text{ref}}\boldsymbol{\theta}_t + 0.15 \cdot \mathbf{a}_t$. 

\subsection{Reward Design}

The reward function is a weighted sum of the imitation reward $r_I$, world frame pose reward $r_\mathcal{W}$, and regularizer $r_R$:
\begin{equation}
    r = 1.0 \cdot \rewardi + 0.5 \cdot \rewardw + 0.1 \cdot \rewardr.
\end{equation}
The imitation reward encourages the robot to follow the reference motion, and is defined as:
\begin{equation}
    \rewardi = r_z \cdot r_\mathrm{v} \cdot r_\mathrm{w} \cdot r_{\phi\theta} \cdot r_{\text{feet}}
\end{equation}
where $r_z$ matches the base height, $r_\mathrm{v}$ matches $xy$ components of base linear velocity, $r_\mathrm{w}$ matches $z$ component of base angular velocities, $r_{\phi\theta}$ matches the roll and pitch angles of the base, and finally $r_{\text{feet}}$ matches the relative foot positions with respect to base. All components are expressed in the robot's base frame.

The world frame pose reward matches the $(x, y)$ position and orientation of the robot's base and that of the reference: $\rewardw = r_{xy} \cdot r_{\mathrm{h}}$. This reward term improves base motion tracking for dynamic movements at high velocities. 

Finally, the regularizer term $\rewardr = r_{\Delta\mathrm{a}} \cdot r_{\mathrm{a}} \cdot r_{\text{slip}}$ consists of penalties on action rate, action value, and foot slip.

We note that the subterms of $\rewardi$, $\rewardw$, and $\rewardr$ are multiplied together. All subterms follow the following form:
\begin{equation}
    r_\mathrm{y} = \exp\left(-\norm{\frac{\hat{\mathbf{y}} - \mathbf{y}}{\sigma_\mathrm{y}}}^2\right), 
    \label{eq:exp_map}
\end{equation}
where $\hat{\mathbf{y}}$ is the desired value of the robot quantity $\mathbf{y}$, and $\sigma_\mathrm{y}$ denotes sensitivity parameter. The detailed reward hyperparameters are listed in~\Cref{tab:reward-hyperparams}. 

\subsection{Other Training Details}

The tracking policy is trained in a physically simulated environment using \emph{NVIDIA IsaacLab} \cite{mittal2023orbit}.
The simulated \emph{Unitree Go2} robot is equipped with a joint PD controller with $K_p=30$ and $K_d=0.5$.
The policy is queried at $\SI{50}{\hertz}$, with each control action executed over four simulation steps.

To facilitate exploration around the reference motion, we implemented reference state initialization~\citep{peng2018deepmimic}. Additionally, we apply domain randomization to improve the general robustness of the tracking policies against the sim-to-real gap and external disturbances. Specifically, we randomize the friction coefficient within the range $[0.5, 1.5]$, vary the mass of each link by $\pm10\%$, perturb the center of mass of each link by up to $\SI{0.05}{\metre}$, and apply random external perturbations to the robot base every $1.5$ to $\SI{2.5}{\second}$. 

The tracking policy is also trained using PPO with the same set of hyperparameters as the motion synthesis policy, except for the entropy coefficient, which is set to $0.002$.


\section{Results}

\begin{figure*}
    \vspace{1mm}
    \centering
    \includegraphics[width=0.92\linewidth]{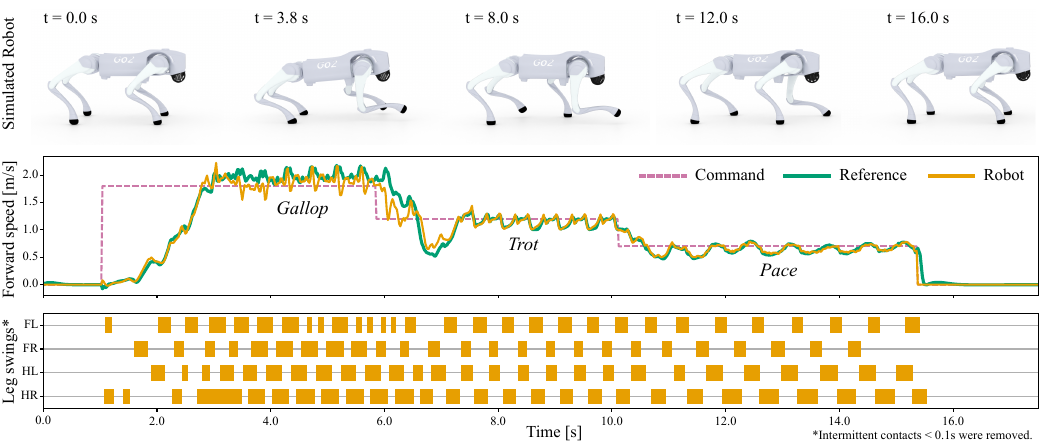}
    \caption{Snapshots of physically simulated \emph{Go2} (\textbf{top}) executing a motion sequence generated by our motion synthesis module in response to varying forward speed commands shown alongside the speed profile (\textbf{middle}) and leg swing timeline (\textbf{bottom}).}
    \label{fig:results-gait-transitions}
    \vspace{-1em}
\end{figure*}

We conducted a series of simulation and real-world experiments to evaluate each component of our framework and to demonstrate the full control pipeline developed using the proposed framework. 
In our experiments, we used a subset of a dog motion dataset \cite{zhang2018mode}, consisting of 13076 pose samples along with their left-right mirrored counterparts. 

\subsection{Evaluation of Kino-dynamic Motion Retargeting}

In the first experiment, we evaluate our kino-dynamic MR strategy against the UVM baseline.
We selected three representative animal motion sequences---\emph{Sit-pace-sit}, \emph{Spin}, and \emph{Gallop}---and retargeted them using both methods, and trained motion tracking policies using five random seeds. The resulting learning curves are shown in \Cref{fig:results-motion-retargeting}(c). 

The evaluation reveals several critical limitations in the UVM baseline. For the \emph{Sit-pace-sit} motion, it produces frequent and excessive limb penetrations, as quantified by the mean penetration value in the bar chart of \Cref{fig:results-motion-retargeting}(b). Consequntly, this leads to persistent ground contact compromises tracking accuracy.
Similarly, for the \emph{Spin} motion, the UVM-retargeted reference exhibits frequent foot slippage, rendering the motion unsuitable for real-world deployment.
Finally, with the high-speed \emph{Gallop} motion---a task where physical feasibility is critical---the UVM baseline's poor tracking performance causes the imitation reward to saturate prematurely. 

By successfully eliminating kinematic artifacts and ensuring dynamic feasibility, our kino-dynamic MR method effectively transforms agile and expressive animal motions into robot-compatible trajectories for reliable hardware execution as shown in \Cref{fig:results-motion-retargeting}(a).

\subsection{Analysis of Motion Synthesis Module}

\begin{figure}[!t]
    \includegraphics[width=0.98\linewidth]{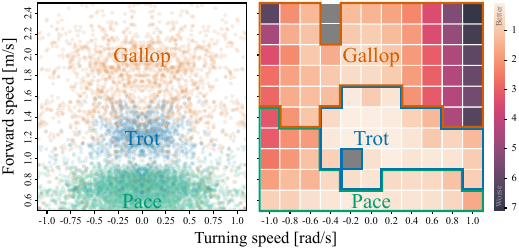}
    \caption{Sample distribution from the motion DB visualized as a scatter plot over forward and turning speeds (\textbf{left}). Command tracking error and corresponding gaits of the motion synthesis module shown as a heatmap. Gray cells indicate failure to transition to suitable gait modes (\textbf{right}).}
    \label{fig:results-mvae-heatmap}
    \vspace{-1em}
\end{figure}

Shifting our focus to motion synthesis, we evaluate the ability of our motion synthesis module to generate motions that accurately follow velocity commands and exhibit appropriate gait transitions. In each trial, the robot starts from a standing pose, and we measure the tracking error over a $\SI{10}{\second}$ period while applying various speed commands.
To generate the heatmap in Figure~\ref{fig:results-mvae-heatmap}, we sweep these commands across a range of $[0.6, 2.4]\, \SI{}{\metre/\second}$ (forward) and $[-1.0, 1.0]\, \SI[]{}{\radian/\second}$ (turning).  The heatmap in \Cref{fig:results-mvae-heatmap} visualizes the weighted sum of mean squared forward speed error $e_{\text{fwd}}$ and turning speed error $e_{\text{turn}}$, computed as $e_{\text{fwd}} + 10 \cdot e_{\text{turn}}$. A higher weight is applied to the turning error to balance the visualization, as its magnitude is relatively small.

The results show that the module generally demonstrates high fidelity in tracking user commands. More importantly, it successfully captures and reproduces the distinct gait patterns present in the dataset.
However, the module exhibits limitations in regions where the training data is sparse, particularly for high-speed galloping commands and during the transition between pacing and trotting. In these cases, the system may disregard user input and instead persist in its previous motion cycle.
We further discuss this limitation in \Cref{sec:conclusion}. 

To better examine the synthesized motion, we generated a motion sequence using our motion synthesis module under varying forward speed commands. 
We then trained a motion tracking policy to follow this sequence and executed it on the physically simulated \emph{Go2}, as shown in \Cref{fig:results-gait-transitions}. 
As the forward speed profile indicates, the reference motion produced by the motion synthesis module responds smoothly and accurately to the commands, transitioning seamlessly from \emph{Gallop} to \emph{Trot} to \emph{Pace} as the speed varies from $\SI{1.8}{\metre/\second}$ to $\SI{1.2}{\metre/\second}$ to $\SI{0.7}{\metre/\second}$.
The tracking policy successfully reproduces this motion on the robot, achieving a root mean squared base velocity error of $\SI{0.11}{\metre/\second}$.

\subsection{Online Motion Synthesis and Control on Hardware}

Finally, we deployed the full control pipeline---comprising the steerable motion synthesis module and the RL tracking controller---on the robot hardware to enable real-time steering. 
For this experiment, the RL motion tracking policy was trained with the motion synthesis module in the loop, which generates reference motion for the policy based on randomly sampled velocity commands. 
As shown in \Cref{fig:hw-experiment-online}, the control pipeline enables the robot to navigate freely with animal-like gait patterns and demonstrates gait switching in response to velocity commands. 
Readers are referred to the supplementary video for comprehensive footage of the experiments\footnote{The video is available at \url{https://youtu.be/DukyUGNYf5A}}.

\section{Conclusion and Future work}
\label{sec:conclusion}

%


This work presents a framework for steerable imitation control that learns multi-modal behaviors emerging from large, unlabeled real-world data in response to user steering commands. Our approach successfully addresses the embodiment gap between the motion source and the target robot, preserve essential motion patterns from raw data, and induces emergent transitions in response to velocity commands.
\Diff{
Notably, our framework requires neither behavior-specific objectives nor specialized guidance to identify behavioral modes. Instead, it autonomously discovers distinct modalities within unlabeled data, mapping these modes and their transitions to user steering commands through a relatively simple reward structure, without the need for explicit scripting. In our experiments, this was demonstrated by the emergence of characteristic gait patterns and fluid transitions as the user adjusts steering inputs.
}

Although the framework is effective in its primary goals, its development also highlighted several key areas for future research.
The primary challenge lies within the steering motion synthesis module, where the module's performance is constrained by the density of the training data, limiting its ability to generalize in sparse data regions. Furthermore, as the module is trained purely kinematically, it can produce motion artifacts---such as overly aggressive movements---at high velocities. A key future direction is to develop synthesis modules that are both robust to data sparsity and physically grounded, eliminating artifacts without sacrificing dynamic range or suffering from the mode collapse seen in some prior physics-aware approaches \cite{peng2021amp, yao2022controlvae, won2022physics}.


Looking ahead, a promising extension of this work involves its application to humanoids, where natural motor skill imitation remains a critical research frontier. Beyond stylistic consistency, we aim to evolve this framework into a comprehensive locomotion pipeline capable of supporting a broader repertoire of skills across challenging obstacles, facilitating seamless transitions that allow robots to adapt fluidly to both complex environmental constraints and dynamic user steering commands.

\addtolength{\textheight}{-12cm}   





\section*{Acknowledgment}

The authors utilized Google Gemini 3 for grammatical polishing and stylistic refinement of the manuscript.

\bibliographystyle{IEEEtranN}
\bibliography{root.bib}


\end{document}